\documentclass[10pt,twocolumn,letterpaper]{article}

\usepackage[pagenumbers]{cvpr} % To force page numbers, e.g. for an arXiv version

\usepackage{graphicx}
\usepackage{amsmath}
\usepackage{amssymb}
\usepackage{booktabs}

\usepackage{multirow}
\usepackage{makecell}
\usepackage[accsupp]{axessibility}  % Improves PDF readability for those with disabilities.

\usepackage[pagebackref,breaklinks,colorlinks]{hyperref}

\usepackage[capitalize]{cleveref}
\crefname{section}{Sec.}{Secs.}
\Crefname{section}{Section}{Sections}
\Crefname{table}{Table}{Tables}
\crefname{table}{Tab.}{Tabs.}

\def\cvprPaperID{5876} % *** Enter the CVPR Paper ID here
\def\confName{CVPR}
\def\confYear{2023}

\begin{document}

\title{ESLAM: Efficient Dense SLAM System Based on Hybrid Representation of Signed Distance Fields}

\author{Mohammad Mahdi Johari\\
Idiap Research Institute, EPFL\\
{\tt\small mohammad.johari@idiap.ch}
\and
Camilla Carta\\
ams OSRAM\\
{\tt\small camilla.carta@ams-osram.com}
\and
François Fleuret\\
University of Geneva, EPFL\\
{\tt\small francois.fleuret@unige.ch}
}

\twocolumn[{
	\maketitle
	\begin{center}
		\vspace{-3ex}
		\captionsetup{type=figure}
		\includegraphics[width=1.0\linewidth]{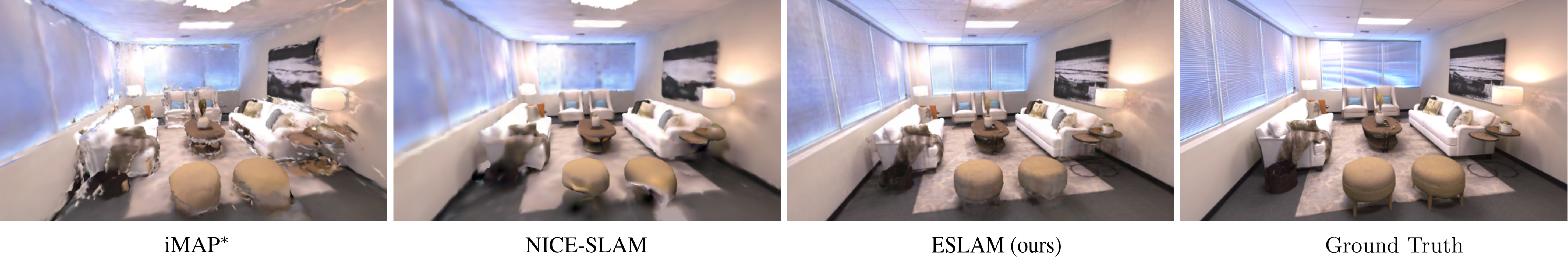}
		\vspace{-4ex}
            \captionof{figure}{Our pre-train-free ESLAM model reconstructs scene details more accurately than existing works: iMAP$^*$~\cite{sucar2021imap} and NICE-SLAM~\cite{zhu2022nice}, while it runs up to $\times$10 faster (see Sec.~\ref{sec:runtime} for runtime analysis). The ground truth image is rendered with ReplicaViewer~\cite{replica19arxiv}.}
    \end{center}
}]

\maketitle

%%%%%%%%% ABSTRACT
\begin{abstract}
   We present ESLAM, an efficient implicit neural representation method for Simultaneous Localization and Mapping (SLAM). ESLAM reads RGB-D frames with unknown camera poses in a sequential manner and incrementally reconstructs the scene representation while estimating the current camera position in the scene. We incorporate the latest advances in Neural Radiance Fields (NeRF) into a SLAM system, resulting in an efficient and accurate dense visual SLAM method. Our scene representation consists of multi-scale axis-aligned perpendicular feature planes and shallow decoders that, for each point in the continuous space, decode the interpolated features into Truncated Signed Distance Field (TSDF) and RGB values. Our extensive experiments on three standard datasets, Replica, ScanNet, and TUM RGB-D show that ESLAM improves the accuracy of 3D reconstruction and camera localization of state-of-the-art dense visual SLAM methods by more than 50\%, while it runs up to $\times$10 faster and does not require any pre-training. Project page: \href{https://www.idiap.ch/paper/eslam}{https://www.idiap.ch/paper/eslam}
\end{abstract}

%%%%%%%%% BODY TEXT
\section{Introduction}
\label{sec:intro}

Dense visual Simultaneous Localization and Mapping (SLAM) is a fundamental challenge in 3D computer vision with several applications such as autonomous driving, robotics, and virtual/augmented reality. It is defined as constructing a 3D map of an unknown environment while simultaneously approximating the camera pose.

While traditional SLAM systems~\cite{mur2017orb, engel2014lsd, newcombe2011dtam, schops2019bad, whelan2015elasticfusion, whelan2012kintinuous} mostly focus on localization accuracy, recent learning-based dense visual SLAM methods~\cite{bloesch2018codeslam, yang2022fd, czarnowski2020deepfactors, sucar2020nodeslam, zhi2019scenecode, teed2021droid, mccormac2017semanticfusion, sunderhauf2017meaningful, tang2018ba, koestler2022tandem} provide meaningful global 3D maps and show reasonable but limited reconstruction accuracy.

Following the advent of Neural Radiance Fields (NeRF)~\cite{mildenhall2020nerf} and the demonstration of their capacity to reason about the geometry of a large-scale scene~\cite{deng2022depth, kosiorek2021nerf, chen2021mvsnerf, johari2022geonerf, wei2021nerfingmvs, jain2021putting, wu2022scalable} and reconstruct 3D surfaces~\cite{yariv2021volume, azinovic2022neural, wang2021neus, sun2022neural, or2022stylesdf, li2022bnv, ortiz2022isdf, zhang2021ners, wang2022neuris}, novel NeRF-based dense SLAM methods have been developed. In particular, iMAP~\cite{sucar2021imap} and NICE-SLAM~\cite{zhu2022nice} utilize neural implicit networks to achieve a consistent geometry representation.

IMAP~\cite{sucar2021imap} represents the geometry with a single huge MLP, similar to NeRF~\cite{mildenhall2020nerf}, and optimizes the camera poses during the rendering process. NICE-SLAM~\cite{zhu2022nice} improves iMAP by storing the representation locally on voxel grids to prevent the forgetting problem. Despite promising reconstruction quality, these methods are computationally demanding for real-time applications, and their ability to capture geometry details is limited. In addition, NICE-SLAM~\cite{zhu2022nice} uses frozen pre-trained MLPs, which limits its generalizability to novel scenes. We take NICE-SLAM~\cite{zhu2022nice} as a baseline and provide the following contributions:
\begin{itemize}
	\item We leverage implicit Truncated Signed Distance Field (TSDF)~\cite{azinovic2022neural} to represent geometry, which converges noticeably faster than the common rendering-based representations like volume density~\cite{sucar2021imap} or occupancy~\cite{zhu2022nice} and results in higher quality reconstruction.
	
	\item Instead of storing features on voxel grids, we propose employing multi-scale axis-aligned feature planes~\cite{chan2022efficient} which leads to reducing the memory footprint growth rate \wrt scene side-length from cubic to quadratic.
	
	\item We benchmark our method on three challenging datasets, Replica~\cite{replica19arxiv}, ScanNet~\cite{dai2017scannet}, and TUM RGB-D~\cite{sturm2012benchmark}, to demonstrate the performance of our method in comparison to existing ones and provide an extensive ablation study to validate our design choices. 
	
\end{itemize}
Thanks to the inherent smoothness of representing the scene with feature planes, our method produces higher-quality smooth surfaces without employing explicit smoothness loss functions like~\cite{wang2022go}.

Concurrent with our work, the followings also propose Radiance Fields-based SLAM systems: iDF-SLAM~\cite{ming2022idf} also uses TSDF, but it is substantially slower and less accurate than NICE-SLAM~\cite{zhu2022nice}. Orbeez-SLAM~\cite{chung2022orbeez} operates in real-time at the cost of poor 3D reconstruction. Compromising accuracy and quality, MeSLAM~\cite{kruzhkov2022meslam} introduces a memory-efficient SLAM. MonoNeuralFusion~\cite{zou2022mononeuralfusion} proposes an incremental 3D reconstruction model, assuming that ground truth camera postures are available. Lastly, NeRF-SLAM~\cite{rosinol2022nerf} presents a monocular SLAM system with hierarchical volumetric Neural Radiance Fields optimized using an uncertainty-based depth loss.

%-------------------------------------------------------------------------
\section{Related Work}

\noindent \textbf{Dense Visual SLAM.} The ubiquity of cameras has made visual SLAM a field of major interest in the last decades. Traditional visual SLAM employs pixel-wise optimization of geometric and/or photometric constraints from image information. Depending on the information source, visual SLAM divides into three main categories: visual-only~\cite{newcombe2011dtam, engel2014lsd, forster2014svo, mur2017orb, Tateno_2017_CVPR, engel2017direct}, visual-inertial~\cite{mourikis2007multi, leutenegger2015keyframe, bloesch2015robust, mur2017visual, qin2018vins, von2018direct} and RGB-D~\cite{newcombe2011kinectfusion, kerl2013dense, endres20133, campos2021orb} SLAM. Visual-only SLAM uses single or multi-camera setups but presents higher technical challenges compared to others. Visual-inertial information can improve accuracy, but complexifies the system and requires an extra calibration step. The advent of the Kinect brought popularity to RGB-D setups with improved performance but had drawbacks such as larger memory and power requirements, and limitation to indoor settings. Recently, learning-based approaches~\cite{bloesch2018codeslam, li2018undeepvo, li2020deepslam, czarnowski2020deepfactors, teed2021droid} have made great advances in the field, improving both accuracy and robustness compared to traditional methods.

\vspace{0.95ex}
\noindent\textbf{Neural Implicit 3D Reconstruction.}  Neural Radiance Fields (NeRF) have impacted 3D Computer Vision applications, such as novel view synthesis~\cite{mildenhall2020nerf, martin2021nerf, verbin2022ref, mildenhall2022nerf}, surface reconstruction~\cite{park2019deepsdf, yariv2021volume, oechsle2021unisurf, wang2021neus, zhang2021ners, wang2022neuris, yariv2020multiview}, dynamic scene representation~\cite{gao2021dynamic, park2021nerfies, pumarola2021d, park2021hypernerf}, and camera pose estimation~\cite{yen2021inerf, wang2021nerf, lin2021barf, jeong2021self, xia2022sinerf}. The exploitation of neural implicit representations for 3D reconstruction at real-world scale is studied in~\cite{azinovic2022neural, wang2022go, bozic2021transformerfusion, choe2021volumefusion, murez2020atlas, sun2021neuralrecon, weder2021neuralfusion, yan2021continual, li2022bnv}. The most related works to ours are iMAP~\cite{sucar2021imap} and NICE-SLAM~\cite{zhu2022nice}. IMAP~\cite{sucar2021imap} presents a NeRF-style dense SLAM system. NICE-SLAM~\cite{zhu2022nice} extends iMAP~\cite{sucar2021imap} by modeling the scene with voxel grid features and decoding them into occupancies using pre-trained MLPs. However, the generalizability of NICE-SLAM~\cite{zhu2022nice} to novel scenes is limited because of the frozen pre-trained MLPs. Another issue is the cubic memory growth rate of their model, which results in using low-resolution voxel grids and losing fine geometry details. In contrast, we employ compact plane-based features~\cite{chan2022efficient} which are directly decoded to TSDF, improving both efficiency and accuracy of localization and reconstruction. \looseness=-1

%-------------------------------------------------------------------------
\section{Method}
\begin{figure*}
    \begin{center}
        \includegraphics[width=0.99\linewidth]{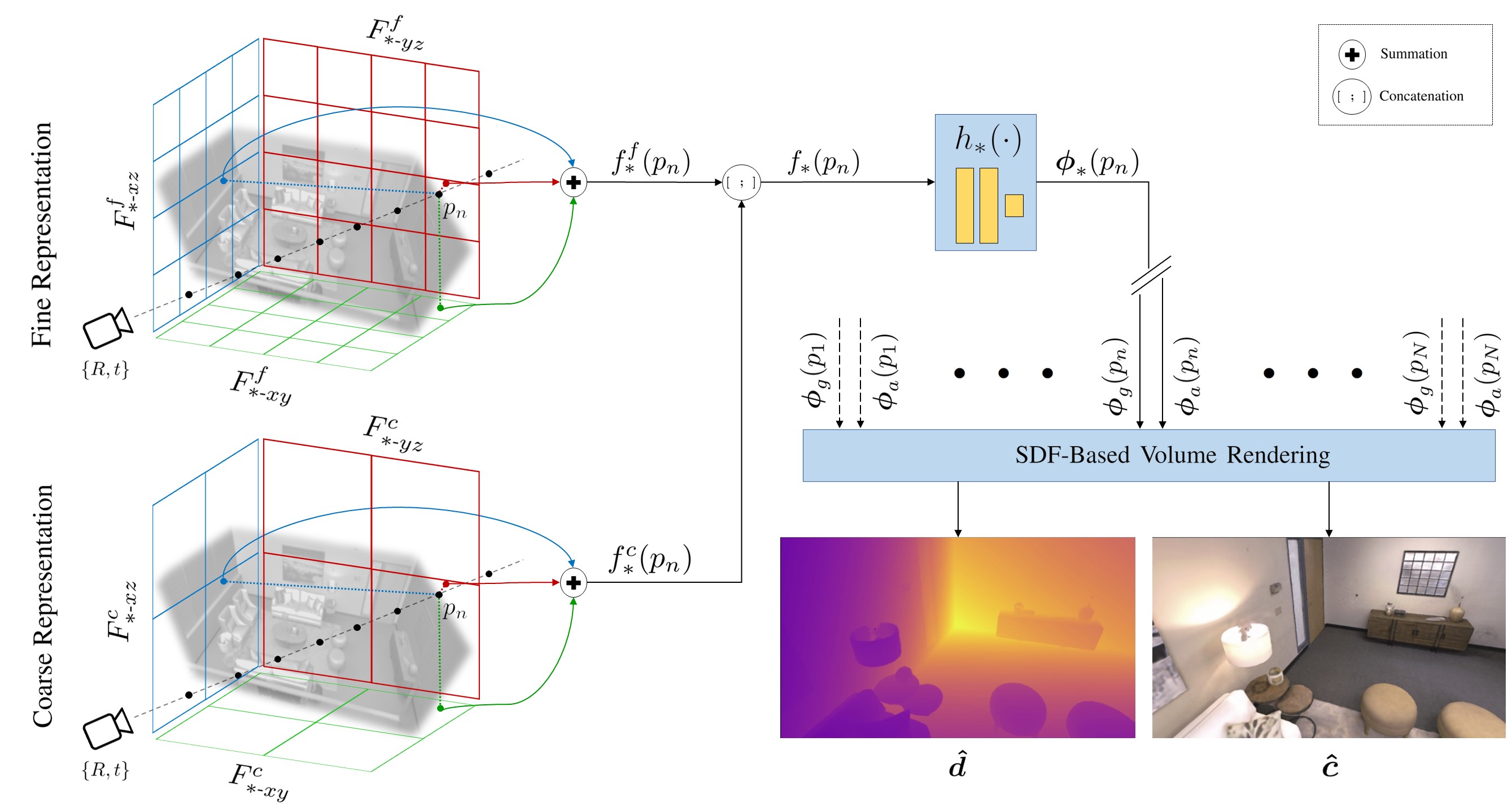}
    \end{center}
    \vspace{-3ex}
    \caption{The overview of ESLAM. Given the symmetry of our processes for geometry and appearance, we exhibit both processes on the same pipeline for simplicity. The symbol~$*$ represents both geometry~$g$ and appearance~$a$, \eg, $f_{*}(p_{n})$ can be either $f_{g}(p_{n})$ or $f_{a}(p_{n})$. At an estimated camera pose~$\{R,t\}$, we cast a ray for each pixel and sample $N$ points~$\{p_{n}\}_{n=1}^{N}$ along it (Sec.~\ref{sec:rendering}). Each point~$p_{n}$ is projected onto the coarse and fine feature planes and bilinearly interpolates the four nearest neighbor features on each plane (Sec.~\ref{sec:feat_planes}). The interpolated features at each level are added together, and the results from both levels are concatenated together to form the inputs~$\{\boldsymbol{f_{g}}(p_{n}), \boldsymbol{f_{a}}(p_{n})\}$ of the decoders~$\{h_{g}, h_{a}\}$ (Sec.~\ref{sec:feat_planes}). The geometry decoder~$h_{g}$ estimates TSDF~$\boldsymbol{\phi_{g}}(p_{n})$ based on $\boldsymbol{f_{g}}(p_{n})$, and the appearance decoder~$h_{a}$ estimates the raw color~$\boldsymbol{\phi_{a}}(p_{n})$ based on $\boldsymbol{f_{a}}(p_{n})$ for each point~$p_{n}$ (Sec.~\ref{sec:feat_planes}). Once TSDFs and raw colors of all points on a ray are generated, our SDF-based rendering process estimates the depth~$\boldsymbol{\hat{d}}$ and the color~$\boldsymbol{\hat{c}}$ for each pixel (Sec.~\ref{sec:rendering}). 
   }
    \label{fig:arch}
    \vspace{-2.0ex}
\end{figure*}

The overview of our method is shown in Fig.~\ref{fig:arch}. Given a set of sequential RGB-D frames~$\{I_{i}, D_{i}\}_{i=1}^M$, our model predicts camera poses~$\{R_{i}| t_{i}\}_{i=1}^M$ and an implicit TSDF~$\boldsymbol{\phi_{g}}$ representation that can be used in marching cubes algorithm~\cite{lorensen1987marching} to extract 3D meshes. We expect TSDF to denote the distance to the closest surface with a positive sign in the free space and a negative sign inside the surfaces. We employ normalized TSDF, such that it is zero on the surfaces and has a magnitude of one at the truncation distance~$T$, which is a hyper-parameter. Sec.~\ref{sec:feat_planes} describes how we represent a scene with axis-aligned feature planes. Sec.~\ref{sec:rendering} walks through the rendering process, which converts raw representations into pixel depths and colors. Sec.~\ref{sec:losses} introduces our loss functions. And Sec.~\ref{sec:map_track} provides the details of localization and reconstruction in our SLAM system.

\subsection{Axis-Aligned Feature Planes} \label{sec:feat_planes}
Although voxel grid-based NeRF architectures~\cite{fridovich2022plenoxels, sun2022direct, wang2022go} exhibit rapid convergence, they struggle with cubical memory growing. Different solutions have been proposed to mitigate the memory growth issue~\cite{muller2022instant, Chen2022ECCV, chan2022efficient}. Inspired by~\cite{chan2022efficient}, we employ a tri-plane architecture (see Fig.~\ref{fig:arch}), in which we store and optimize features on perpendicular axis-aligned planes. Mimicking the trend in voxel-based methods~\cite{muller2022instant, sun2022direct, Chen2022ECCV}, we propose using feature planes at two scales, \ie coarse and fine. Coarse-level representation allows efficient reconstruction of free space with fewer sample points and optimization iterations. Moreover, we suggest employing separate feature planes for representing geometry and appearance, which mitigates the forgetting problem for geometry reconstruction since appearance fluctuates more frequently in a scene than geometry.

Specifically, we use three coarse feature planes~\{$F^{c}_{g\text{-}xy}$, $F^{c}_{g\text{-}xz}$, $F^{c}_{g\text{-}yz}$\} and three fine ones~\{$F^{f}_{g\text{-}xy}$, $F^{f}_{g\text{-}xz}$, $F^{f}_{g\text{-}yz}$\} for representing the geometry. Similarly, three coarse~\{$F^{c}_{a\text{-}xy}$, $F^{c}_{a\text{-}xz}$, $F^{c}_{a\text{-}yz}$\} and three fine~\{$F^{f}_{a\text{-}xy}$, $F^{f}_{a\text{-}xz}$, $F^{f}_{a\text{-}yz}$\} planes are used for representing appearance of a scene. This architecture prevents model size from growing cubically with the scene side-length as is the case for voxel-based models.

To reason about the geometry of a point~$p$ in the continuous space, we first project it onto all the geometry planes. The geometry feature~$\boldsymbol{f_{g}}(p)$ for point~$p$ is then formed by 1) bilinearly interpolating the four nearest neighbors on each feature plane, 2) summing the interpolated coarse features and the fine ones respectively into the coarse output~$f^{c}_{g}(p)$ and fine output~$f^{f}_{g}(p)$, and 3) concatenating the outputs together. Formally: 
\begin{align}
	f^{c}_{g}(p) &= F^{c}_{g\text{-}xy}(p) + F^{c}_{g\text{-}xz}(p) + F^{c}_{g\text{-}yz}(p) \notag \\
	f^{f}_{g}(p) &= F^{f}_{g\text{-}xy}(p) + F^{f}_{g\text{-}xz}(p) + F^{f}_{g\text{-}yz}(p) \notag \\
	\boldsymbol{f_{g}}(p) &= [f^{c}_{g}(p) ; f^{f}_{g}(p)]
\end{align}

The appearance feature~$\boldsymbol{f_{a}}(p)$ is obtained similarly:
\begin{align}
	f^{c}_{a}(p) &= F^{c}_{a\text{-}xy}(p) + F^{c}_{a\text{-}xz}(p) + F^{c}_{a\text{-}yz}(p) \notag \\
	f^{f}_{a}(p) &= F^{f}_{a\text{-}xy}(p) + F^{f}_{a\text{-}xz}(p) + F^{f}_{a\text{-}yz}(p) \notag \\
	\boldsymbol{f_{a}}(p) &= [f^{c}_{a}(p) ; f^{f}_{a}(p)]
\end{align}

These features are decoded into TSDF~$\boldsymbol{\phi_{g}}(p)$ and raw color~$\boldsymbol{\phi_{a}}(p)$ values via shallow two-layer MLPs~\{$h_{g}$, $h_{a}$\}:
\begin{align}
	\boldsymbol{\phi_{g}}(p) = h_{g}\left(\boldsymbol{f_{g}}(p)\right) \text{~~and~~} \boldsymbol{\phi_{a}}(p) = h_{a}\left(\boldsymbol{f_{a}}(p)\right)
\end{align}

These raw TSDF and color outputs can be utilized for depth/color rendering as well as mesh extraction.

\subsection{SDF-Based Volume Rendering} \label{sec:rendering}
When processing input frame~$i$, emulating the ray casting in NeRF~\cite{mildenhall2020nerf}, we select random pixels and calculate their corresponding rays using the current estimate of the camera pose~$\{R_{i}| t_{i}\}$. For rendering the depths and colors of the rays, we first sample $N_{strat}$ samples on each ray by stratified sampling and then sample additional $N_{imp}$ points near surfaces. For pixels with ground truth depths, the $N_{imp}$ additional points are sampled uniformly inside the truncation distance~$T$ \wrt the depth measurement, whereas for other pixels, $N_{imp}$ points are sampled with the importance sampling technique~\cite{mildenhall2020nerf, martin2021nerf, wang2022go, sucar2021imap} based on the weights computed for the stratified samples.

For all $N=N_{strat}+N_{imp}$ points on a ray~$\{p_{n}\}_{n=1}^{N}$, we query TSDF~$\boldsymbol{\phi_{g}}(p_{n})$ and raw color~$\boldsymbol{\phi_{a}}(p_{n})$ from our networks and use the SDF-Based rendering approach in StyleSDF~\cite{or2022stylesdf} to convert SDF values to volume densities:
\begin{align}
	\boldsymbol{\sigma}(p_{n}) &= \beta \cdot \text{Sigmoid} \left(-\beta \cdot \boldsymbol{\phi_{g}}(p_{n}) \right)
\end{align}
where $\beta$ is a learnable parameter that controls the sharpness of the surface boundary. Negative values of SDF push Sigmoid toward one, resulting in volume density inside the surface. The volume density then is used for rendering the color and depth of each ray:
\begin{align}
    w_{n} &= \exp \left( -\sum_{k=1}^{n-1} \boldsymbol{\sigma}(p_{k}) \right) \left( 1 - \exp \left(-\boldsymbol{\sigma}\left(p_{n}\right) \right)\right) \notag 
    \\
	\boldsymbol{\hat{c}} &= \sum_{n=1}^{N} w_{n} \boldsymbol{\phi_{a}}(p_{n}) \text{\phantom{00} and \phantom{00}} \boldsymbol{\hat{d}} = \sum_{n=1}^{N} w_{n} z_{n}
\end{align}
where $z_{n}$ is the depth of point~$p_{n}$ \wrt the camera pose.

\subsection{Loss Functions} \label{sec:losses}
One advantage of TSDF over other representations, such as occupancy, is that it allows us to use per-point losses, along with rendering ones. These losses account for the rapid convergence of our model. Following the practice in~\cite{azinovic2022neural}, assuming a batch of rays~$R$ with ground truth depths are selected, we define the free space loss as:
\begin{align}
	\mathcal{L}_{fs} &= \frac{1}{|R|} \sum_{r \in R} \frac{1}{|P_{r}^{fs}|} \sum_{p \in P_{r}^{fs}} (\boldsymbol{\phi_{g}}(p) - 1)^2
\end{align}
where $P_{r}^{fs}$ is a set of points on the ray~$r$ that lie between the camera center and the truncation region of the surface measured by the depth sensor. This loss function encourages TSDF~$\boldsymbol{\phi_{g}}$ to have a value of one in the free space.

For sample points close to the surface and within the truncation region, we use the signed distance objective, which leverages the depth sensor measurement to approximate the signed distance field:
\begin{align}
    &\mathcal{L}_{T}(P_{r}^{T}) = \notag \\
    &\frac{1}{|R|} \sum_{r \in R} \frac{1}{|P_{r}^{T}|} \sum_{p \in P_{r}^{T}} \left( z(p) + \boldsymbol{\phi_{g}}(p) \cdot T - D(r) \right)^2
\end{align}
where $z(p)$ is the planar depth of point~$p$ \wrt camera, $T$ is the truncation distance, $D(r)$ is the ray depth measured by the sensor, and $P_{r}^{T}$ is a set of points on the ray~$r$ that lie in the truncation region, \ie $|z(p) - D(r)| < T$. We apply the same loss to all points in the truncation region, but we differentiate the importance of points that are closer to the surface in the middle of the truncation region~$P_{r}^{T\text{-}m}$ from those that are at the tail of the truncation region~$P_{r}^{T\text{-}t}$. Formally, we define $P_{r}^{T\text{-}m}$ as a set of points that $|z(p) - D(r)| < 0.4T$, and define $P_{r}^{T\text{-}t} = P_{r}^{T} - P_{r}^{T\text{-}m}$, then:
\begin{align}
	\mathcal{L}_{T\text{-}m} &= \mathcal{L}_{T}(P_{r}^{T\text{-}m}) \text{\phantom{00} and \phantom{00}} \mathcal{L}_{T\text{-}t} = \mathcal{L}_{T}(P_{r}^{T\text{-}t})
\end{align}

This enables us to decrease the importance of $\mathcal{L}_{T\text{-}t}$ in mapping, which leads to having a smaller effective truncation distance, reducing artifacts in occluded areas, and reconstructing with higher accuracy while leveraging the entire truncation distance in camera tracking. 

In addition to these two per-point loss functions, we also employ reconstruction losses. For pixels with ground truth depths, we impose consistency between the rendered depth and the depth measured by the sensor:
\begin{align}
	\mathcal{L}_{d} &= \frac{1}{|R|} \sum_{r \in R} \left( \boldsymbol{\hat{d}}(r) - D(r) \right)^2
\end{align}

Similarly, we impose consistency between the pixel colors and rendered colors:
\begin{align}
	\mathcal{L}_{c} &= \frac{1}{|R|} \sum_{r \in R} \left( \boldsymbol{\hat{c}}(r) - I(r) \right)^2
\end{align}
where $I(r)$ is the pixel color of ray~$r$.

The global loss function of our method is defined as:
\begin{align}
	\hspace{-0.75em} \mathcal{L} &= \lambda_{fs}\mathcal{L}_{fs} + \lambda_{T\text{-}m}\mathcal{L}_{T\text{-}m} + \lambda_{T\text{-}t}\mathcal{L}_{T\text{-}t} + \lambda_{d}\mathcal{L}_{d} + \lambda_{c}\mathcal{L}_{c}
\end{align}
where $\{\lambda_{fs}, \lambda_{T\text{-}m}, \lambda_{T\text{-}t}, \lambda_{d}, \lambda_{c}\}$ are the weighting coefficients. Note that $\mathcal{L}_{c}$ is defined on all rays in a training batch, while other losses are only imposed on rays with ground truth measured depths. The global objective is the same for both mapping and tracking in our method, but the weighting coefficients are different.

\begin{table*}[!t]
    \begin{center}
        \begin{tabular}{l|cccc|cc}
            \Xhline{2\arrayrulewidth}
            \multirow{2}{*}{Method} & \multicolumn{4}{c|}{Reconstruction (cm)} &  \multicolumn{2}{c}{Localization (cm)}  \\
            & \small Depth L1$\downarrow$ & \small Acc.$\downarrow$ & \small Comp.$\downarrow$ & \small Comp. Ratio (\%)$\uparrow$ & \small ATE Mean$\downarrow$ & \small ATE RMSE$\downarrow$ \\
            
            \hline
            iMAP$^{*}$~\cite{sucar2021imap} & 8.23 $\pm$ 0.88 & 7.16 $\pm$ 0.26 & 5.83 $\pm$ 0.27 & 67.17 $\pm$ 2.70 & 2.59 $\pm$ 0.58 & 3.42 $\pm$ 0.87 \\
            NICE-SLAM~\cite{zhu2022nice} & 3.29 $\pm$ 0.33 & 1.66 $\pm$ 0.07 & 1.63 $\pm$ 0.05 & 96.74 $\pm$ 0.36 & 1.56 $\pm$ 0.29 & 2.05 $\pm$ 0.45 \\
            ESLAM (ours) & \textbf{1.18 $\pm$ 0.05} & \textbf{0.97 $\pm$ 0.02} & \textbf{1.05 $\pm$ 0.01} & \textbf{98.60 $\pm$ 0.07} & \textbf{0.52 $\pm$ 0.03} & \textbf{0.63 $\pm$ 0.05} \\
            
            \Xhline{2\arrayrulewidth}
        \end{tabular}
    \end{center}
    \vspace{-3ex}
    \caption{Quantitative comparison of our proposed ESLAM with existing NeRF-based dense visual SLAM models on the Replica dataset~\cite{replica19arxiv} for both reconstruction and localization accuracy. The results are the average and standard deviation of five runs on eight scenes of the Replica dataset~\cite{replica19arxiv}. Our method outperforms previous works by a high margin and has lower variances, indicating it is also more stable from run to run. The evaluation metrics for reconstruction are L1 loss (cm) between rendered and ground truth depth maps of 1000 random camera poses, reconstruction accuracy (cm), reconstruction completion (cm), and completion ratio (\%). The evaluation metrics for localization are mean and RMSE of ATE (cm)~\cite{sturm2012benchmark}. For the details of the evaluations for each scene, refer to the supplementary. It should also be noted that our method runs up to $\times$10 faster on this dataset (see Sec.~\ref{sec:runtime} for runtime analysis).}
    \label{table:quantitative_replica}
    % \vspace{-1ex}
\end{table*}

\begin{figure*}
    \begin{center}
        \includegraphics[width=0.99\linewidth]{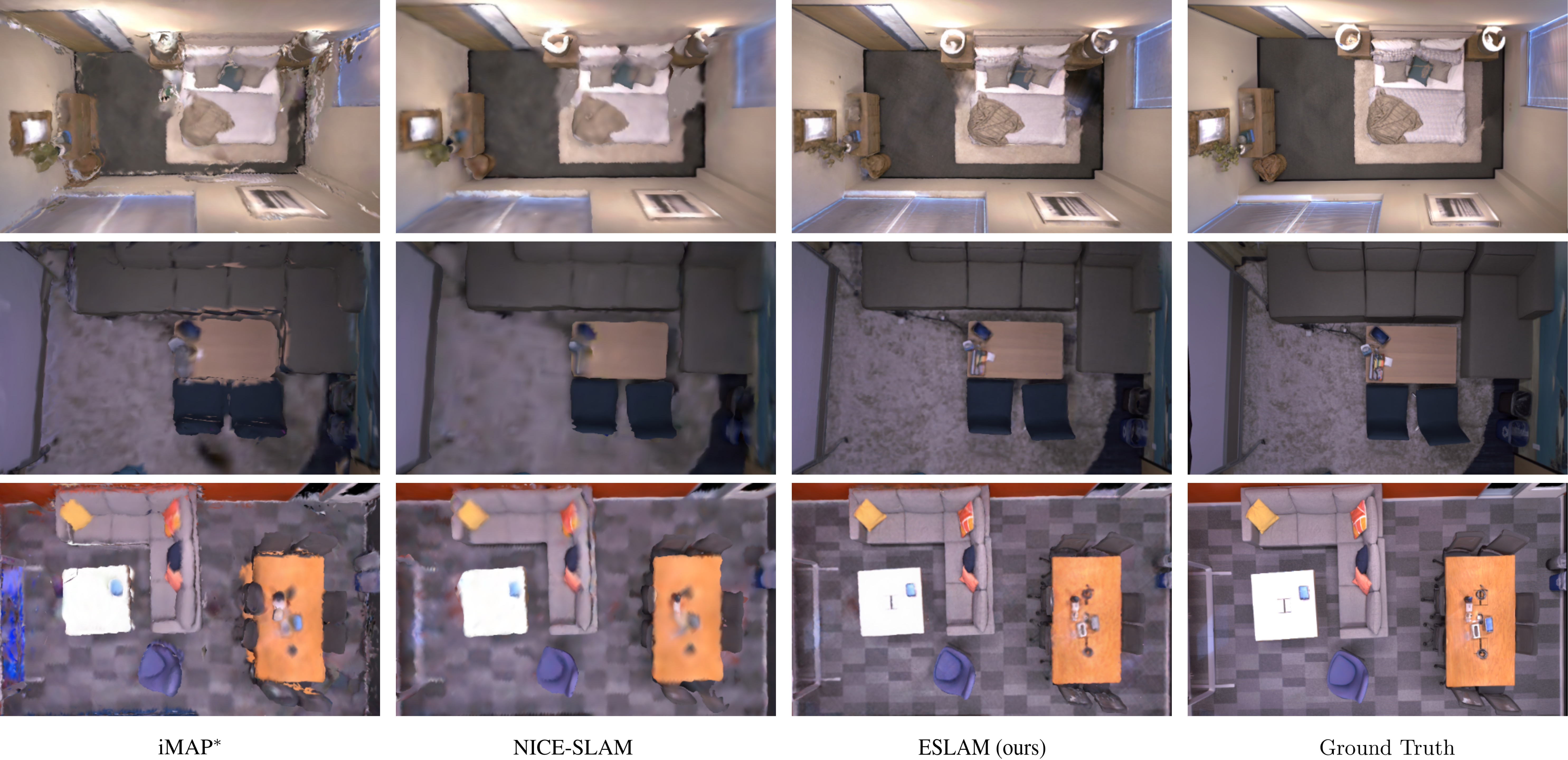}
    \end{center}
    \vspace{-3ex}
    \caption{Qualitative comparison of our proposed ESLAM method's geometry reconstruction with existing NeRF-based dense visual SLAM models, iMAP$^*$~\cite{sucar2021imap} and NICE-SLAM~\cite{zhu2022nice}, on the Replica dataset~\cite{replica19arxiv}. Our method produces more accurate detailed geometry as well as higher-quality textures. The ground truth images are rendered with the ReplicaViewer software~\cite{replica19arxiv}. It should also be noted that our method runs up to $\times$10 faster on this dataset (see Sec.~\ref{sec:runtime} for runtime analysis). For further qualitative analysis on this dataset, as well as videos demonstrating the localization and reconstruction process, refer to the supplementary.}
    \label{fig:qualitative_replica}
    \vspace{-2ex}
\end{figure*}

\subsection{Mapping and Tracking} \label{sec:map_track}
\noindent\textbf{Mapping.} Our scene representation, \ie the feature planes and MLP decoders, are randomly initialized at the beginning. With the first input frame~$\{I_{0}, D_{0}\}$, we fix the camera pose and optimize the feature planes and MLP decoders to best represent the first frame. For subsequent inputs, we update the scene representation iteratively every $k$ frames, and add the latest frame to the global keyframe list, following the practice in iMAP~\cite{sucar2021imap} and NICE-SLAM~\cite{zhu2022nice}. For mapping, we first choose $|R|$ pixels randomly from $W$ frames, which include the current frame, the previous two keyframes, and $W - 3$ frames randomly selected from the keyframe list. Then, we jointly optimize the feature planes, MLP decoders, and camera poses of the $W$ selected frames using the loss functions introduced in Sec.~\ref{sec:losses}. Unlike NICE-SLAM~\cite{zhu2022nice}, our method does not require a staged-optimization policy, and we simply optimize all scene parameters and camera poses simultaneously.

\vspace{1ex}
\noindent\textbf{Tracking.} The localization process of our method is initiated for each input frame. The current estimate of the camera parameters, represented by translation vectors and quaternion rotations\cite{shoemake1985animating}~$\{R|t\}$, are optimized solely based on our global loss function (see Sec.~\ref{sec:losses}) with the gradient-based Adam optimizer~\cite{adam}. No second-order optimizers or manifold operations are employed for camera tracking in our method. We exclude rays with no ground truth depths and outlier pixels from each optimization step. A pixel is considered an outlier if the difference between its measured depth and rendered depth is ten times greater than the batch's median rendered depth error.

\section{Experiments}
In this section, we validate that our method outperforms existing implicit representation-based methods in both localization and reconstruction accuracy on three standard benchmarks while running up to $\times$10 faster.

\begin{table*}[!t]
    \begin{center}
        \begin{tabular}{ll|cccccc|c}
            \Xhline{2\arrayrulewidth}
            Method & ATE & Sc. 0000 & Sc. 0059 & Sc. 0106 & Sc. 0169 & Sc. 0181 & Sc. 0207 & Ave.\\
            
            \hline
            iMAP$^{*}$~\cite{sucar2021imap} & Mean & \small 34.2 $\pm$ 12.8 & \small 13.0 $\pm$ 2.4 & \small 12.9 $\pm$ 1.7 & \small 33.6 $\pm$ 15.3 & \small 20.8 $\pm$ 3.8 & \small 18.6 $\pm$ 6.0 & \small 22.2 $\pm$ 7.0 \\
            & RMSE & \small 42.7 $\pm$ 16.6 & \small 17.8 $\pm$ 7.4 & \small 15.0 $\pm$ 1.7 & \small 39.1 $\pm$ 18.2 & \small 24.7 $\pm$ 5.8 & \small 20.1 $\pm$ 6.8 & \small 26.6 $\pm$ 9.4 \\
            
            \hline
            NICE-SLAM~\cite{zhu2022nice} & Mean & \small \phantom{0}9.9 $\pm$ \phantom{0}0.4 & \small 11.9 $\pm$ 1.8 & \small \phantom{0}7.0 $\pm$ 0.2 & \small \phantom{0}9.2 $\pm$ \phantom{0}1.0 & \small 12.2 $\pm$ 0.3 & \small \phantom{0}5.5 $\pm$ 0.3 & \small \phantom{0}9.3 $\pm$ 0.7 \\
            & RMSE & \small 12.0 $\pm$ \phantom{0}0.5 & \small 14.0 $\pm$ 1.8 & \small \phantom{0}7.9 $\pm$ 0.2 & \small 10.9 $\pm$ \phantom{0}1.1 & \small 13.4 $\pm$ 0.3 & \small \phantom{0}6.2 $\pm$ 0.4 & \small 10.7 $\pm$ 0.7 \\
            
            \hline
            ESLAM (ours) & Mean & \small \textbf{\phantom{0}6.5 $\pm$ \phantom{0}0.1} & \small \textbf{\phantom{0}6.4 $\pm$ 0.4} & \small \textbf{\phantom{0}6.7 $\pm$ 0.1} & \small \textbf{\phantom{0}5.9 $\pm$ \phantom{0}0.1} & \small \textbf{\phantom{0}8.3 $\pm$ 0.2} & \small \textbf{\phantom{0}5.4 $\pm$ 0.1} & \small \textbf{\phantom{0}6.5 $\pm$ 0.2} \\
            & RMSE & \small \textbf{\phantom{0}7.3 $\pm$ \phantom{0}0.2} & \small \textbf{\phantom{0}8.5 $\pm$ 0.5} & \small \textbf{\phantom{0}7.5 $\pm$ 0.1} & \small \textbf{\phantom{0}6.5 $\pm$ \phantom{0}0.1} & \small \textbf{\phantom{0}9.0 $\pm$ 0.2} & \small \textbf{\phantom{0}5.7 $\pm$ 0.1} & \small \textbf{\phantom{0}7.4 $\pm$ 0.2} \\
            
            \Xhline{2\arrayrulewidth}
        \end{tabular}
    \end{center}
    \vspace{-3ex}
    \caption{Quantitative comparison of our proposed ESLAM method's localization accuracy with existing NeRF-based dense visual SLAM models on the ScanNet dataset~\cite{dai2017scannet}. The results are the average and standard deviation of five runs on each scene of ScanNet~\cite{dai2017scannet}. Our method outperforms previous works and has lower variances, indicating it is also more stable from run to run. The evaluation metrics for localization are mean and RMSE of ATE (cm)~\cite{sturm2012benchmark}. It should also be noted that our method runs up to $\times$6 faster on this dataset (see Sec.~\ref{sec:runtime} for runtime analysis).}
    \label{table:quantitative_scannet}
    % \vspace{-1.0ex}
\end{table*}

\begin{figure*}
    \begin{center}
        \includegraphics[width=0.99\linewidth]{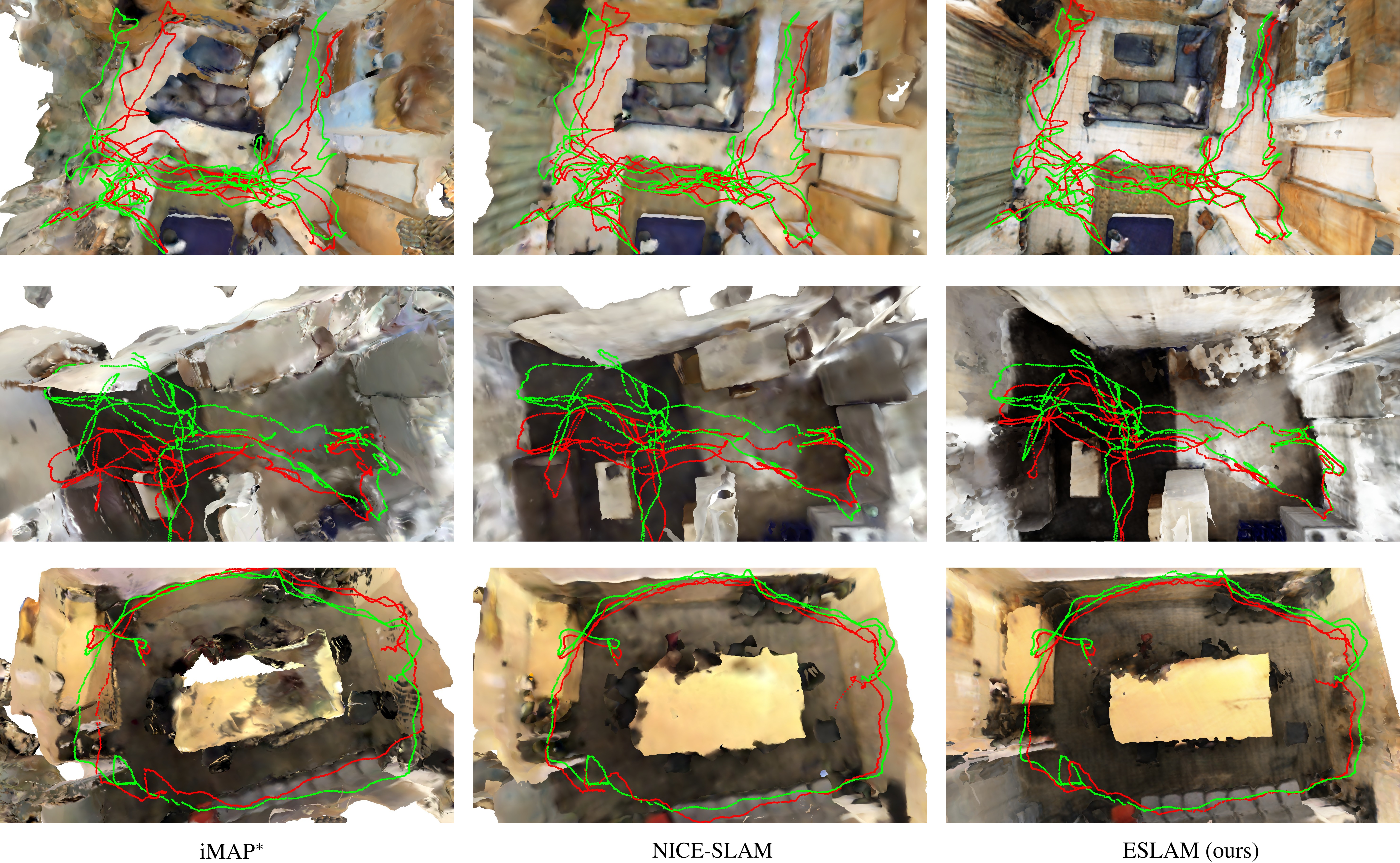}
    \end{center}
    \vspace{-3ex}
    \caption{Qualitative comparison of our proposed ESLAM method's localization accuracy with existing NeRF-based dense visual SLAM models, iMAP$^*$~\cite{sucar2021imap} and NICE-SLAM~\cite{zhu2022nice}, on the ScanNet dataset~\cite{dai2017scannet}. The ground truth camera trajectory is shown in \textcolor{green}{green}, and the estimated trajectory is shown in \textcolor{red}{red}. Our method predicts more accurate camera trajectories and does not suffer from drifting issues. It should also be noted that our method runs up to $\times$6 faster on this dataset (see Sec.~\ref{sec:runtime} for runtime analysis).}
    \label{fig:qualitative_tracking}
    \vspace{-2.5ex}
\end{figure*}

\vspace{1ex}
\noindent\textbf{Baselines.} We compare our method to two existing state-of-the-art NeRF-based dense visual SLAM methods: iMAP~\cite{sucar2021imap} and NICE-SLAM~\cite{zhu2022nice}. Because iMAP is not open source, we use the iMAP$^*$ model in our experiment, which is the reimplementation of iMAP in~\cite{zhu2022nice}.

\vspace{1ex}
\noindent\textbf{Datasets.} We evaluate our method on three standard 3D benchmarks: Replica~\cite{replica19arxiv}, ScanNet~\cite{dai2017scannet}, and TUM RGB-D~\cite{sturm2012benchmark} datasets. We select the same scenes for evaluation as NICE-SLAM~\cite{zhu2022nice}.

\vspace{1ex}
\noindent\textbf{Metrics.} We borrow our evaluation metrics from NICE-SLAM~\cite{zhu2022nice}. For evaluating scene geometry, we use both 2D and 3D metrics. For the 2D metric, we render depth maps from 1000 random camera poses in each scene and calculate the L1 difference between depths from ground truth meshes and the reconstructed ones. For the 3D metrics, we consider reconstruction accuracy~[cm], reconstruction completion~[cm], and completion ratio~[$<$~5~cm \%]. For evaluating these metrics, we build a TSDF volume for a scene with a resolution of 1 cm and use the marching cubes algorithm~\cite{lorensen1987marching} to obtain scene meshes. Before evaluating the 3D metrics for our method and for the baselines, we perform mesh culling as recommended in~\cite{azinovic2022neural, wang2022go}. For this purpose, we remove faces from a mesh that are not inside any camera frustum or are occluded in all RGB-D frames. For evaluating camera localization, we use ATE~\cite{sturm2012benchmark}.

\begin{figure*}
    \begin{center}
        \includegraphics[width=0.99\linewidth]{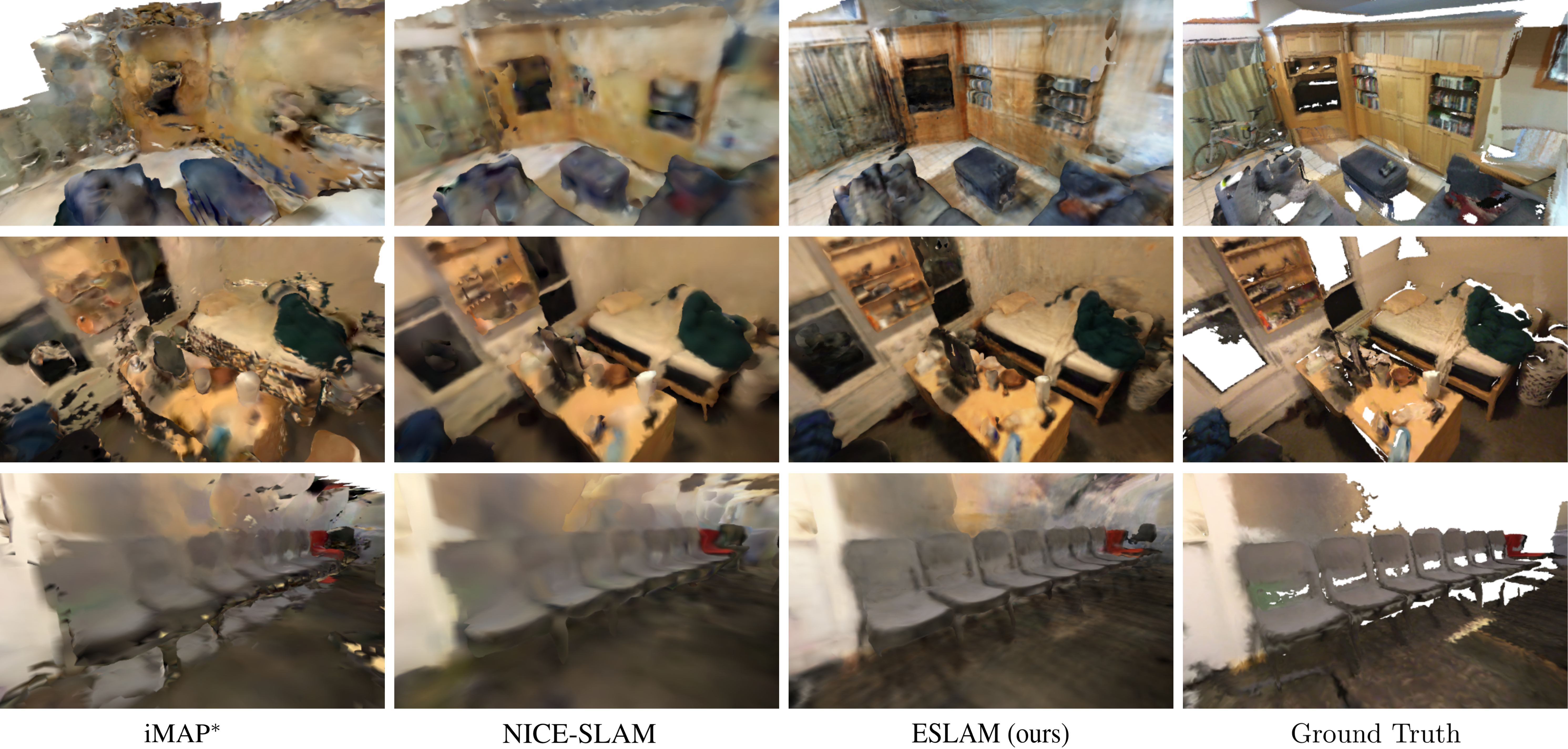}
    \end{center}
    \vspace{-3ex}
    \caption{Qualitative comparison of our proposed ESLAM method's geometry reconstruction with existing NeRF-based dense visual SLAM models, iMAP$^*$~\cite{sucar2021imap} and NICE-SLAM~\cite{zhu2022nice}, on the ScanNet dataset~\cite{dai2017scannet}. Our method produces more accurate detailed geometry as well as higher-quality textures. The appearance of white backgrounds in ground truth meshes is due to the fact that the ground truth meshes of the ScanNet dataset~\cite{dai2017scannet} are incomplete. It should also be noted that our method runs up to $\times$6 faster on this dataset (see Sec.~\ref{sec:runtime} for runtime analysis).}
    \label{fig:qualitative_scannet}
    \vspace{-3ex}
\end{figure*}

\vspace{1ex}
\noindent\textbf{Implementation Details.} The truncation distance~$T$ is set to 6 cm in our method. We employ coarse feature planes with a resolution of 24 cm for both geometry and appearance. For fine feature planes, we use a resolution of 6 cm for geometry and 3 cm for appearance. All feature planes have 32 channels, resulting in a 64-channel concatenated feature input for the decoders. The decoders are two-layer MLPs with 32 channels in the hidden layer. For the Replica~\cite{replica19arxiv} dataset, we sample $N_{strat}=32$ points for stratified sampling and $N_{imp}=8$ points for importance sampling on each ray. And for the ScanNet~\cite{dai2017scannet} and TUM RGB-D~\cite{sturm2012benchmark} datasets, we set $N_{strat}=48$ and $N_{imp}=8$.

We use different set of loss coefficients for mapping and tracking. During mapping we set $\lambda_{fs}=5$, $\lambda_{T\text{-}m}=200$, $\lambda_{T\text{-}t}=10$, $\lambda_{d}=0.1$, and $\lambda_{c}=5$. And during tracking, we set $\lambda_{fs}=10$, $\lambda_{T\text{-}m}=200$, $\lambda_{T\text{-}t}=50$, $\lambda_{d}=1$, and $\lambda_{c}=5$. These coefficients are obtained by performing grid search in our experiments. For further details of our implementation, refer to the supplementary.

\subsection{Experimental Results}
\noindent\textbf{Evaluation on Replica~\cite{replica19arxiv}.} We provide the quantitative analysis of our experimental results on eight scenes of the Replica dataset~\cite{replica19arxiv} in Tab.~\ref{table:quantitative_replica}. The numbers represent the average and standard deviation of the metrics for five independent runs. As shown in Tab.~\ref{table:quantitative_replica}, our method outperforms the baselines for both reconstruction and localization accuracy. Our method also has lower variances, indicating that it is more stable and more robust than existing methods.

Qualitative analysis on the Replica dataset~\cite{replica19arxiv} is provided in Fig.~\ref{fig:qualitative_replica}. The results show that our method reconstructs the details of the scenes more accurately and produces fewer artifacts. Although it is not the focus of this paper, our method also produces higher-quality colors for the reconstructed meshes.

\vspace{1ex}
\noindent\textbf{Evaluation on ScanNet~\cite{dai2017scannet}.} We also benchmark ours and existing methods on multiple large scenes from ScanNet~\cite{dai2017scannet} to evaluate and compare their scalability. For evaluating camera localization, we conduct five independent experiments on each scene and report the average and standard deviation of the mean and RMSE of ATE~\cite{sturm2012benchmark} in Tab.~\ref{table:quantitative_scannet}. As demonstrated in the table, our method's localization is more accurate than existing methods. Our method is also considerably more stable from run to run as it has much lower standard deviations. We provide qualitative analysis of camera localization, along with geometry reconstruction, in Fig.~\ref{fig:qualitative_tracking}. The results show that our method does not suffer from any large drifting and is more robust than existing methods.

Since the ground truth meshes of the ScanNet dataset~\cite{dai2017scannet} are incomplete, we only provide qualitative analysis for geometry reconstruction, similar to previous works. The qualitative comparison in Fig.~\ref{fig:qualitative_scannet} validates that our model reconstructs more precise geometry and detailed textures compared to existing approaches.

\begin{figure*}
    \begin{center}
        \includegraphics[width=0.99\linewidth]{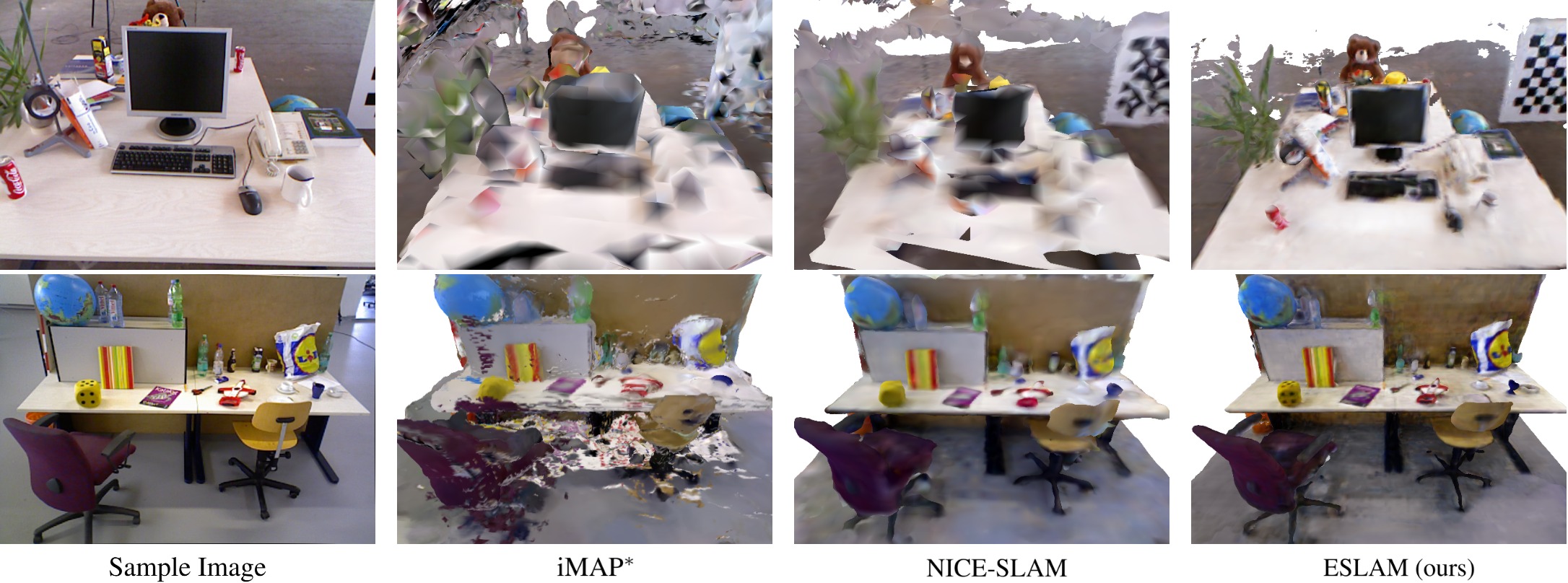}
    \end{center}
    \vspace{-3ex}
    \caption{Qualitative comparison of our proposed ESLAM method's geometry reconstruction with existing NeRF-based dense visual SLAM models, iMAP$^*$~\cite{sucar2021imap} and NICE-SLAM~\cite{zhu2022nice}, on the TUM RGB-D dataset~\cite{sturm2012benchmark}. Our method produces more accurate detailed geometry as well as higher-quality textures. Since there are no ground truth meshes for this dataset, we depict a sample input image.}
    \label{fig:qualitative_tum}
    \vspace{-2ex}
\end{figure*}

\begin{table}[!t]
    \begin{center}
        \begin{tabular}{l|ccc}
            \Xhline{2\arrayrulewidth}
            & fr1/desk & fr2/xyz & fr3/office \\

            \hline
            iMAP$^{*}$~\cite{sucar2021imap} & 4.90 & 2.05 & 5.80 \\
            NICE-SLAM~\cite{zhu2022nice} & 2.85 & 2.39 & 3.02 \\
            ESLAM (ours) & \textbf{2.47} & \textbf{1.11} & \textbf{2.42} \\ 
            
            \Xhline{2\arrayrulewidth}
        \end{tabular}
    \end{center}
    \vspace{-3ex}
    \caption{Quantitative comparison of our proposed ESLAM method's localization accuracy with existing NeRF-based dense visual SLAM models on the TUM RGB-D dataset~\cite{sturm2012benchmark}. The evaluation metric is ATE RMSE$\downarrow$ (cm)~\cite{sturm2012benchmark}.}
    \label{table:tum}
    \vspace{-1ex}
\end{table}

\vspace{1ex}
\noindent\textbf{Evaluation on TUM RGB-D~\cite{sturm2012benchmark}.} To further contrast the robustness of our method with the existing ones, we conduct an evaluation study on the real-world TUM RGB-D dataset~\cite{sturm2012benchmark}. Since there are no ground truth meshes for the scenes in this dataset, we only present the localization results in Tab.~\ref{table:tum} and the qualitative analysis of rendered meshes in Fig.~\ref{fig:qualitative_tum}.

\begin{table}[!t]    
    \begin{center}
        \begin{tabular}{l|l|c|cl}
            \Xhline{2\arrayrulewidth}
            & \multirow{2}{*}{\small Method} & \small Speed & \multicolumn{2}{c}{\small Memory} \\
            &  & \small FPT (s) & \small \# Param. & \small Grow. R. \\
            
            \hline
            \multirow{3}{*}{\rotatebox[origin=c]{90}{\small Replica}} & \small iMAP$^{*}$~\cite{sucar2021imap} &  5.20 & \textbf{\phantom{0}0.22 M} & ? \\
            & \small NICE-SLAM~\cite{zhu2022nice} & 2.10 & 12.18 M & $O(L^3)$ \\
            & \small ESLAM (ours) & \textbf{0.18} & \phantom{0}6.79 M & $O(L^2)$ \\

            \hline
            \multirow{3}{*}{\rotatebox[origin=c]{90}{\small ScanNet}} & \small iMAP$^{*}$~\cite{sucar2021imap} &  5.20 & \textbf{\phantom{0}0.22 M} & ? \\
            & \small NICE-SLAM~\cite{zhu2022nice} & 3.35 & 22.04 M & $O(L^3)$ \\
            & \small ESLAM (ours) & \textbf{0.55} & 17.63 M & $O(L^2)$ \\
            
            \Xhline{2\arrayrulewidth}
        \end{tabular}
    \end{center}
    \vspace{-3ex}
    \caption{Runtime analysis of our method in comparison with existing ones in terms of average frame processing time (FPT), number of parameters, and model size growth rate \wrt scene side-length~$L$. All methods are benchmarked with an NVIDIA GeForce RTX 3090 GPU on \texttt{room0} of Replica~\cite{replica19arxiv} and \texttt{scene0000} of ScanNet~\cite{dai2017scannet}. Our method is significantly faster and does not grow cubically in size \wrt scene side-length~$L$. Note that iMAP~\cite{sucar2021imap} represents a whole scene in a single MLP, hence its small number of parameters. Accordingly, the scalability and growth rate of iMAP~\cite{sucar2021imap} \wrt the scene side-length~$L$ are also unclear.}
    \label{table:runtime}
    \vspace{-3.0ex}
\end{table}

\subsection{Runtime Analysis} \label{sec:runtime}
We evaluate the speed and size of our method in comparison with existing approaches in Tab.~\ref{table:runtime}. We report the average frame processing time (FPT), the number of parameters of the model, and memory growth rate \wrt scene side-length for the scenes \texttt{room0} of Replica~\cite{replica19arxiv} and \texttt{scene0000} of ScanNet~\cite{dai2017scannet} datasets. All methods are benchmarked with an NVIDIA GeForce RTX 3090 GPU. The results indicate that our method is significantly faster than previous works on both datasets. Furthermore, in contrast to NICE-SLAM~\cite{zhu2022nice}, our model size is smaller and does not grow cubically with the scene side-length.

\section{Conclusion}
We presented ESLAM, a dense visual SLAM approach that leverages the latest advances in the Neural Radiance Fields study to improve both the speed and accuracy of neural implicit-based SLAM systems. We proposed replacing the voxel grid representation with axis-aligned feature planes to prevent the model size from growing cubically with respect to the scene side-length. We also demonstrated that modeling the scene geometry with Truncated Signed Distance Field (TSDF) leads to efficient and high-quality surface reconstruction. We verified through extensive experiments that our approach outperforms existing methods significantly in both reconstruction and localization accuracy while running up to one order of magnitude faster.

ESLAM accepts and deals with the forgetting problem in exchange for memory preservation. Due to the structure of our feature plane representation, updating features to adapt to new geometry may affect previously reconstructed geometries. To address this issue, we keep track of previous keyframes and allocate a large portion of computation resources to retain and remember previously reconstructed regions. Although ESLAM is substantially faster than competing approaches, handling the forgetting problem more efficiently could further reduce frame processing time.

\bigbreak\noindent\textbf{Acknowledgement} \\
This research was supported by ams OSRAM.

%%%%%%%%% REFERENCES
{\small
\bibliographystyle{ieee_fullname}
\bibliography{egbib}
}

\end{document}

\title{Supplementary Materials for \\ ESLAM: Efficient Dense SLAM System Based on Hybrid Representation of Signed Distance Fields}

\author{Mohammad Mahdi Johari\\
Idiap Research Institute, EPFL\\
{\tt\small mohammad.johari@idiap.ch}
\and
Camilla Carta\\
ams OSRAM\\
{\tt\small camilla.carta@ams-osram.com}
\and
François Fleuret\\
University of Geneva, EPFL\\
{\tt\small francois.fleuret@unige.ch}
}

\maketitle
\iftoggle{cvprpagenumbers}{}{%
     \thispagestyle{empty}
   }

%%%%%%%%% BODY TEXT
\section{Further Implementation Details}
This section provides additional implementation details of our method. For the sake of completeness, we also reiterate the points mentioned in the main article.

The truncation distance~$T$ is set to 6 cm in our method. We employ coarse feature planes with a resolution of 24 cm for both geometry and appearance. For fine feature planes, we use a resolution of 6 cm for geometry and 3 cm for appearance. All feature planes have 32 channels, resulting in a 64-channel concatenated feature input for the decoders. The decoders are two-layer MLPs with 32 channels in the hidden layer. ReLU activation function is used for the hidden layer, and Tanh and Sigmoid are respectively used for the output layers of TSDF and raw colors.

We use different set of loss coefficients for mapping and tracking. During mapping we set $\lambda_{fs}=5$, $\lambda_{T\text{-}m}=200$, $\lambda_{T\text{-}t}=10$, $\lambda_{d}=0.1$, and $\lambda_{c}=5$. And during tracking, we set $\lambda_{fs}=10$, $\lambda_{T\text{-}m}=200$, $\lambda_{T\text{-}t}=50$, $\lambda_{d}=1$, and $\lambda_{c}=5$. These coefficients are obtained by performing grid search in our experiments.

For the scenes from Replica~\cite{replica19arxiv}, we sample $N_{strat}=32$ points for stratified sampling and $N_{imp}=8$ points for importance sampling on each ray. We perform 15 optimization iterations for mapping and randomly select 4000 rays for each iteration. For camera tracking, 2000 rays are chosen at random and 8 optimization iterations are performed. Since ScanNet's~\cite{dai2017scannet} scenes are at a larger scale and more challenging, we set $N_{strat}=48$ and $N_{imp}=8$. Also, we perform 30 optimization iterations for both mapping and tracking in ScanNet's~\cite{dai2017scannet} scenes. For the scenes in TUM RGB-D dataset~\cite{sturm2012benchmark}, we similarly set $N_{strat}=48$ and $N_{imp}=8$. For this dataset, We perform 60 optimization iterations for mapping and 200 optimization iterations for tracking, and we randomly sample 5000 rays for each iteration.

We initiate the mapping process every 4 input frames and use a window of $W=20$ keyframes for jointly optimizing the feature planes, MLP decoders, and camera poses of the selected keyframes. We use Adam~\cite{adam} for optimizing all learnable parameters of our method and set the learning rates according to a simple grid search in our experiments. We use a learning rate of 0.001 for the MLP decoders and a learning rate of 0.005 for the feature planes. We always use a learning rate of 0.001 for the camera poses, \ie rotation and translation $\{R,t\}$, of the selected keyframes during the joint optimization of the mapping step. During the tracking step in the Replica's~\cite{replica19arxiv} scenes, we use a learning rate of 0.001 for camera rotation and translation. For camera tracking in the scenes of ScanNet~\cite{dai2017scannet}, we use a learning rate of 0.0005 for camera translation and a learning rate of 0.0025 for camera rotation. Lastly, For camera tracking in the scenes of TUM RGB-D~\cite{sturm2012benchmark}, we use a learning rate of 0.01 for camera translation and a learning rate of 0.002 for camera rotation. We model the camera rotation parameters with quaternions~\cite{shoemake1985animating}.

Once all input frames are processed, and for evaluation purposes, we build a TSDF volume for each scene and use the marching cubes algorithm~\cite{lorensen1987marching} to obtain 3D meshes. We do not employ any post-processing for our representation or extracted meshes except that for quantitative evaluation, we cull faces from a mesh that are not inside any camera frustum or are occluded in all RGB-D frames. To ensure fairness, we do the same mesh culling before evaluating the previous approaches.

\section{Ablation Study} \label{sec:ablation}
In this section, we conduct various experiments to show the robustness of our method in different experimental settings and to validate our architecture design choices.

\vspace{1ex}
\noindent\textbf{Robustness to Depth Quality.} In this experiment, we evaluate how robust the methods are to the quality of input depths. Accordingly, we downsample input depths of \texttt{room0} of the replica dataset~\cite{replica19arxiv} to $\frac{1}{8}$ of the original resolution. The results in Tab.~\ref{table:ab_depth} reveal that our method's reconstruction and localization are less sensitive to the resolution of input depth maps.

\begin{table}[!t]
    \begin{center}
        \begin{tabular}{l|l|ccl}
            \Xhline{2\arrayrulewidth}
            & \small Method & \small ATE$\downarrow$ & \small Acc.$\downarrow$ & \small Comp.$\downarrow$ \\
            
            \hline
            \multirow{2}{*}{\small $\frac{1}{1}$ D}
            & \small NICE-SLAM~\cite{zhu2022nice} & 1.69 & 1.71 & 1.69 \\
            & \small ESLAM (ours) & \textbf{0.71} & \textbf{1.07} & \textbf{1.12} \\

            \hline
            \multirow{2}{*}{\small $\frac{1}{8}$ D} & \small NICE-SLAM~\cite{zhu2022nice} & 2.01 & 2.18 & 1.98 \\
            & \small ESLAM (ours) & \textbf{0.72} & \textbf{1.16} & \textbf{1.23} \\
            
            \Xhline{2\arrayrulewidth}
        \end{tabular}
    \end{center}
    \vspace{-3ex}
    \caption{Robustness to depth resolution comparison of our method with NICE-SLAM~\cite{zhu2022nice} in terms of ATE RMSE (cm), reconstruction accuracy (cm), and reconstruction completion (cm) on \texttt{room0} of the Replica~\cite{replica19arxiv} dataset. Our method's accuracy is less affected when input depth is downsampled by a factor of $\frac{1}{8}$.}
    \label{table:ab_depth}
    \vspace{-0.0ex}
\end{table}

\vspace{1ex}
\noindent\textbf{Keyframe Policy.} Whenever we perform a mapping step for an input frame, we always include that frame in our global keyframe list (see Sec.~3.4 in the main paper). NICE-SLAM~\cite{zhu2022nice}, on the other hand, only updates its keyframe list once per 10 mapping steps. To make sure that our evaluations are fair, we also run NICE-SLAM~\cite{zhu2022nice} with our own keyframe updating policy on \texttt{room0} of the Replica~\cite{replica19arxiv} dataset. The results in Tab.~\ref{table:ab_key} show that NICE-SLAM~\cite{zhu2022nice} only slightly benefits from this updating policy.

\begin{table}[!t]
    \begin{center}
        \begin{tabular}{l|ccl}
            \Xhline{2\arrayrulewidth}
            \small Method & \small ATE$\downarrow$ & \small Acc.$\downarrow$ & \small Comp.$\downarrow$ \\

            \hline
            \small NICE-SLAM~\cite{zhu2022nice} & 1.69 & 1.71 & 1.69 \\
            \small NICE-SLAM w/ Our Key. Policy & 1.65 & 1.68 & 1.66 \\

            \Xhline{2\arrayrulewidth}
        \end{tabular}
    \end{center}
    \vspace{-3ex}
    \caption{Analysis of the impact of our keyframe updating policy on NICE-SLAM~\cite{zhu2022nice} (Sec.~3.4 in the main paper). The experiment is conducted on \texttt{room0} of Replica~\cite{replica19arxiv}, and the metrics are ATE RMSE (cm), reconstruction accuracy (cm), and reconstruction completion (cm). NICE-SLAM~\cite{zhu2022nice} only slightly benefits from our updating policy.}
    \label{table:ab_key}
    \vspace{-2ex}
\end{table}

\vspace{1ex}
\noindent\textbf{Our Design Choices.} We conduct multiple experiments in Tab.~\ref{table:ab_choices} to defend our design choices in ESLAM. These experiments are conducted on the scenes in the Replica~\cite{replica19arxiv} and ScanNet~\cite{dai2017scannet} datasets, and the details of the experimental settings are as follows. (a)~We use shared feature planes for geometry and appearance (see Sec.~3.1 and Fig.~2 in the main paper). (b)~We only employ coarse feature planes (see Sec.~3.1 and Fig.~2 in the main paper). (c)~We only employ fine feature planes (see Sec.~3.1 and Fig.~2 in the main paper). (d)~We add the interpolated coarse~$f^{c}_{*}(p_{n})$ and fine~$f^{f}_{*}(p_{n})$ features instead of concatenating them (see Sec.~3.1 and Fig.~2 in the main paper). (e)~We discard importance sampling and use stratified sampling for all $N$ points on a ray (see Sec.~3.2 in the main paper). (f)~We only exploit depth inputs and discard color rendering and RGB inputs (see Sec.~3.2 in the main paper). (g)~We do not consider separate loss functions for the points that are at the tail of the truncation region~$P_{r}^{T\text{-}t}$ and for the points that are in the middle~$P_{r}^{T\text{-}m}$ (see Sec.~3.3 in the main paper). (h)~We do not jointly optimize camera poses during the mapping step (see Sec.~3.4 in the main paper). (i)~We evaluate our full model. Note that due to the incompleteness of ScanNet's~\cite{dai2017scannet} ground truth meshes, we only evaluate localization accuracy on this dataset.

\section{Additional Qualitative Analysis}
\begin{figure*}
    \begin{center}
        \includegraphics[width=0.97\linewidth]{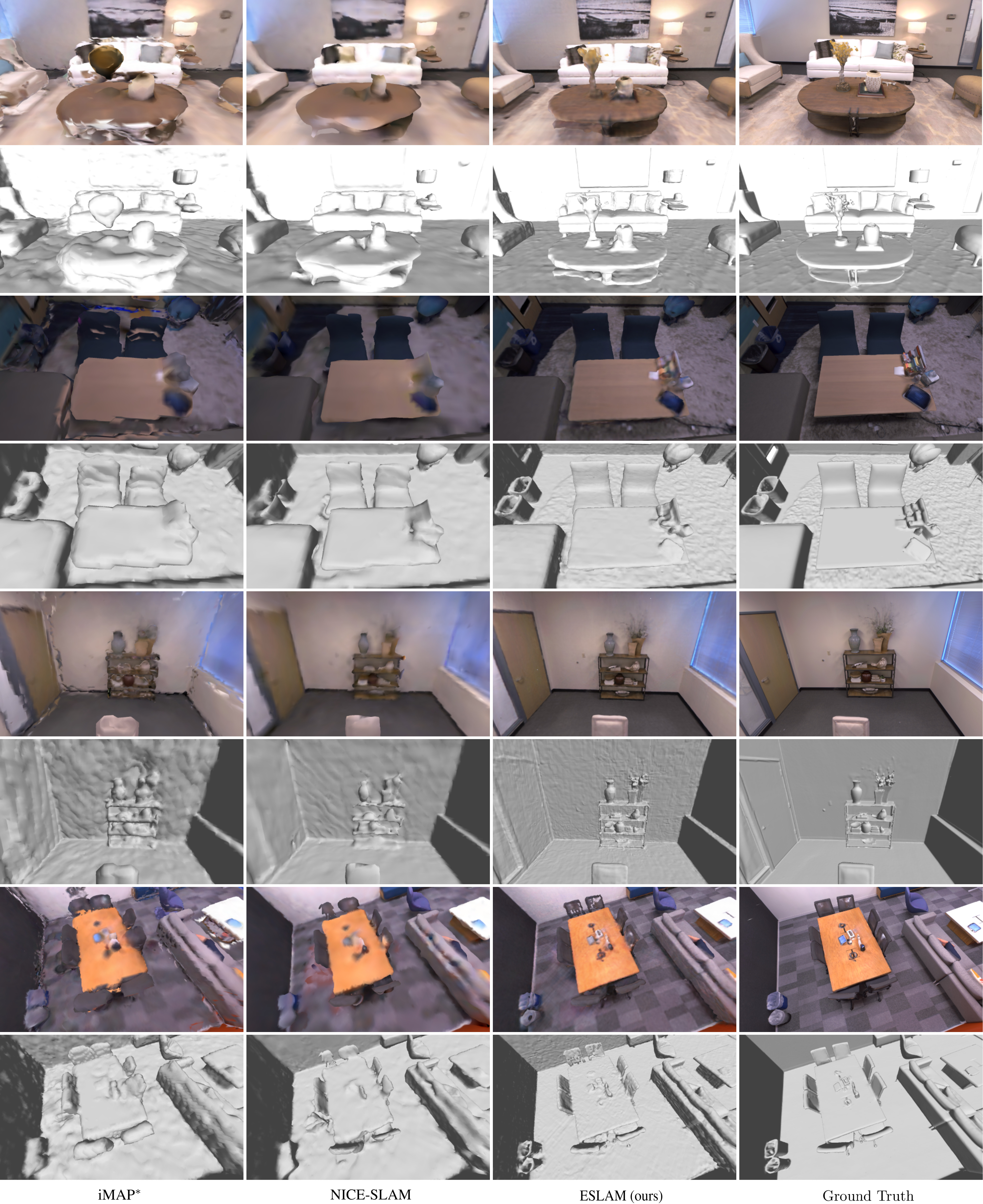}
    \end{center}
    \vspace{-3.0ex}
    \caption{Qualitative comparison of our method's scene reconstruction with iMAP$^*$~\cite{sucar2021imap} and NICE-SLAM~\cite{zhu2022nice} on Replica~\cite{replica19arxiv}. Our method produces more accurate detailed geometry as well as higher-quality textures. The scenes are rendered with both textured and untextured meshes and the ground truth textured images are rendered with the ReplicaViewer software~\cite{replica19arxiv}. It should also be noted that our method runs up to $\times$10 faster on this dataset (see Sec.~4.2 in the main paper for runtime analysis).}
    \label{fig:supp}
\end{figure*}

This section provides additional qualitative analysis to contrast the capability of our method to preserve scene details in comparison to previous NeRF-based dense visual SLAM methods, iMAP$^*$~\cite{sucar2021imap} and NICE-SLAM~\cite{zhu2022nice}. We provide this analysis on the Replica dataset~\cite{replica19arxiv} in Fig.~\ref{fig:supp} with both textured and untextured meshes. The results demonstrate that our method produces more accurate meshes with fewer artifacts.

\section{Per-Scene Breakdown of the Results}
In this section, we breakdown the quantitative analysis of Tab.~1 in the main paper into a per-scene analysis. Tab.~\ref{table:per_scene} shows the per-scene quantitative evaluation of our method in comparison with iMAP$^*$~\cite{sucar2021imap} and NICE-SLAM~\cite{zhu2022nice} on the Replica dataset~\cite{replica19arxiv}. As it is shown in Tab.~\ref{table:per_scene}, our method outperforms previous approaches in all scenes of Replica~\cite{replica19arxiv}. Also, lower variances in our experiments are an indication that our method is more stable from run to run. \looseness=-1

\section{Effect of Frame Processing Time} \label{sec:iters}
In this section, we investigate the trade-off between frame processing time and our method's reconstruction and localization accuracy. In this study, we increase the number of optimization iterations during the mapping and tracking. By default, our ESLAM method performs $Iter_{m}=15$ optimization iterations during mapping and $Iter_{t}=8$ optimization iterations during tracking for the scenes of the Replica dataset~\cite{replica19arxiv}. We define ESLAM~x2 as our method when we double the number of optimization iterations, \ie $Iter_{m}=30$ and $Iter_{t}=16$. And similarly, we define ESLAM~x10 as our method with $Iter_{m}=150$ and $Iter_{t}=80$.

Tab.~\ref{table:iters} provides a quantitative analysis of ESLAM~x2 and ESLAM~x10, as well as a comparison with our default ESLAM method and existing approaches. The results show that at the cost of increased frame processing time, our method yields more accurate scene reconstruction and camera trajectory. It should be noted that even ESLAM~x10 runs faster than the existing state-of-the-art method, NICE-SLAM~\cite{zhu2022nice}.

\begin{figure*}[!t]
    \begin{center}
        \includegraphics[width=1.0\linewidth]{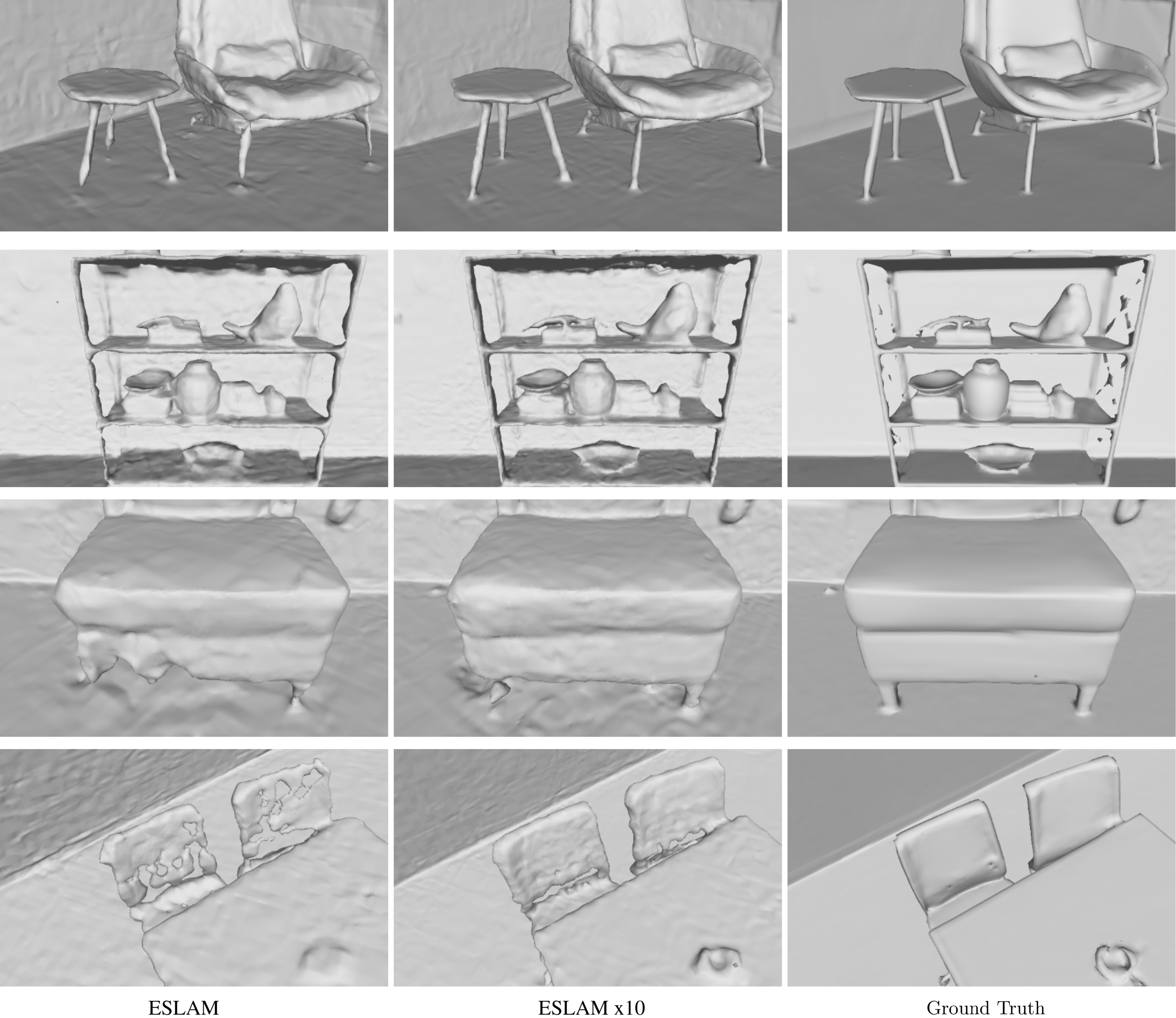}
    \end{center}
    \vspace{-3.0ex}
    \caption{Qualitative analysis of the effect of the number of optimization iterations during mapping and tracking on our method's reconstruction quality. ESLAM~x10 is our method when we multiply the number of optimization iterations by 10. Refer to Sec.~\ref{sec:iters} for the details of this experiment, and see Tab.~\ref{table:iters} for the quantitative analysis.}
    \label{fig:iters}
\end{figure*}

Fig.~\ref{fig:iters} provides a qualitative analysis of ESLAM~x10 compared to our default ESLAM method. In this analysis, we render the scenes with untextured meshes to contrast the quality of geometry reconstruction. While the quality difference is subtle, Fig.~\ref{fig:iters} indicates that increasing the number of optimization iterations results in more accurate geometry reconstruction and smoother surfaces.

\clearpage

\begin{table*}[!t]
    \begin{center}
        \begin{tabular}{l|c|ccc}
            \Xhline{2\arrayrulewidth}
            \multirow{2}{*}{Experiment} & \multicolumn{1}{c|}{ScanNet~\cite{dai2017scannet}} &  \multicolumn{3}{c}{Replica~\cite{replica19arxiv}}  \\
            
            & ATE$\downarrow$ & ~~~ATE$\downarrow$~~~ & Accuracy$\downarrow$ & Compeletion$\downarrow$ \\
    
            \hline
            a. Using shared feature planes for geometry and appearance. & \phantom{0}7.49 & 0.65 & 0.99 & 1.08 \\
            b. Using only the coarse planes and discarding the fine ones. & \phantom{0}7.53 & 0.97 & 1.12 & 1.29 \\
            c. Using only the fine planes and discarding the coarse ones. & \phantom{0}8.27 & 0.72 & 1.00 & 1.09 \\
            d. Replacing the concatenation with a summation. & \phantom{0}7.55 & 0.64 & 0.98 & 1.07 \\
            e. No importance sampling. & \phantom{0}7.44 & 0.67 & 1.08 & 1.14 \\
            f. No color rendering. & \phantom{0}8.31 & 0.68 & 1.03 & 1.08 \\
            g. One loss function for the whole truncation region. & \phantom{0}8.28 & 0.71 & 1.01 & 1.10 \\
            h. No camera pose optimization during mapping. & 11.27 & 4.85 & 2.23 & 2.21 \\
            i. Full ESLAM method. & \phantom{0}\textbf{7.38} & \textbf{0.63} & \textbf{0.97} & \textbf{1.05} \\
    
            \Xhline{2\arrayrulewidth}
        \end{tabular}
    \end{center}
    \vspace{-3ex}
    \caption{Ablation study of our design choices on the ScanNet~\cite{dai2017scannet} and Replica~\cite{replica19arxiv} datasets. The metrics are ATE RMSE (cm), reconstruction accuracy (cm), and reconstruction completion (cm). For the details of this study, see Sec.~\ref{sec:ablation}.}
    \label{table:ab_choices}
    \vspace{-2.0ex}
\end{table*}

\begin{table*}[!t]
    \begin{center}
            \begin{tabular}{l|l|cccc|cc}
            \Xhline{2\arrayrulewidth}
            & \multirow{2}{*}{Methods} & \multicolumn{4}{c|}{Reconstruction (cm)} &  \multicolumn{2}{c}{Localization (cm)}  \\
            &  & \small Depth L1$\downarrow$ & \small Acc.$\downarrow$ & \small Comp.$\downarrow$ & \small Comp. Ratio (\%)$\uparrow$ & \small ATE Mean$\downarrow$ & \small ATE RMSE$\downarrow$ \\
            
            \hline
            \multirow{3}{*}{\rotatebox[origin=c]{90}{room0}} & 
            iMAP$^{*}$~\cite{sucar2021imap} & \phantom{0}6.56 $\pm$ 0.39 & \phantom{0}5.89 $\pm$ 0.19 & 6.07 $\pm$ 0.22 & 66.55 $\pm$ 1.58 & 3.12 $\pm$ 0.84 & 5.23 $\pm$ 1.41 \\
            & NICE-SLAM~\cite{zhu2022nice} & \phantom{0}2.77 $\pm$ 0.13 & \phantom{0}1.71 $\pm$ 0.03 & 1.69 $\pm$ 0.03 & 97.61 $\pm$ 0.09 & 1.43 $\pm$ 0.09 & 1.69 $\pm$ 0.17 \\
            & ESLAM (Ours) & \textbf{\phantom{0}0.97 $\pm$ 0.04} & \textbf{\phantom{0}1.07 $\pm$ 0.01} & \textbf{1.12 $\pm$ 0.01} & \textbf{99.06 $\pm$ 0.05} & \textbf{0.61 $\pm$ 0.06} & \textbf{0.71 $\pm$ 0.13} \\

            \hline
            \multirow{3}{*}{\rotatebox[origin=c]{90}{room1}} & 
            iMAP$^{*}$~\cite{sucar2021imap} & \phantom{0}5.97 $\pm$ 1.14 & \phantom{0}5.71 $\pm$ 0.31 & 5.57 $\pm$ 0.40 & 66.04 $\pm$ 3.45 & 2.54 $\pm$ 0.37 & 3.09 $\pm$ 0.48 \\
            & NICE-SLAM~\cite{zhu2022nice} & \phantom{0}2.52 $\pm$ 0.11 & \phantom{0}1.36 $\pm$ 0.03 & 1.34 $\pm$ 0.04 & 98.60 $\pm$ 0.14 & 1.70 $\pm$ 0.29 & 2.13 $\pm$ 0.24 \\
            & ESLAM (Ours) & \textbf{\phantom{0}1.07 $\pm$ 0.07} & \textbf{\phantom{0}0.85 $\pm$ 0.01} & \textbf{0.88 $\pm$ 0.01} & \textbf{99.64 $\pm$ 0.06} & \textbf{0.56 $\pm$ 0.02} & \textbf{0.70 $\pm$ 0.02} \\

            \hline
            \multirow{3}{*}{\rotatebox[origin=c]{90}{room2}} & 
            iMAP$^{*}$~\cite{sucar2021imap} & \phantom{0}7.82 $\pm$ 0.94 & \phantom{0}6.34 $\pm$ 0.32 & 5.47 $\pm$ 0.27 & 69.87 $\pm$ 4.15 & 2.31 $\pm$ 0.20 & 2.58 $\pm$ 0.19 \\
            & NICE-SLAM~\cite{zhu2022nice} & \phantom{0}3.54 $\pm$ 0.35 & \phantom{0}1.75 $\pm$ 0.06 & 1.71 $\pm$ 0.03 & 96.52 $\pm$ 0.26 & 1.41 $\pm$ 0.24 & 1.87 $\pm$ 0.39 \\
            & ESLAM (Ours) & \textbf{\phantom{0}1.28 $\pm$ 0.07} & \textbf{\phantom{0}0.93 $\pm$ 0.01} & \textbf{1.05 $\pm$ 0.01} & \textbf{98.84 $\pm$ 0.06} & \textbf{0.43 $\pm$ 0.01} & \textbf{0.52 $\pm$ 0.01} \\

            \hline
            \multirow{3}{*}{\rotatebox[origin=c]{90}{office0}} & 
            iMAP$^{*}$~\cite{sucar2021imap} & \phantom{0}7.57 $\pm$ 0.70 & \phantom{0}7.44 $\pm$ 0.26 & 5.13 $\pm$ 0.37 & 70.97 $\pm$ 3.52 & 1.69 $\pm$ 1.06 & 2.40 $\pm$ 1.05 \\
            & NICE-SLAM~\cite{zhu2022nice} & \phantom{0}2.17 $\pm$ 0.14 & \phantom{0}1.43 $\pm$ 0.06 & 1.56 $\pm$ 0.05 & 96.30 $\pm$ 0.33 & 1.12 $\pm$ 0.22 & 1.26 $\pm$ 0.24 \\
            & ESLAM (Ours) & \textbf{\phantom{0}0.86 $\pm$ 0.02} & \textbf{\phantom{0}0.85 $\pm$ 0.01} & \textbf{0.96 $\pm$ 0.01} & \textbf{98.34 $\pm$ 0.05} & \textbf{0.42 $\pm$ 0.03} & \textbf{0.57 $\pm$ 0.04} \\

            \hline
            \multirow{3}{*}{\rotatebox[origin=c]{90}{office1}} & 
            iMAP$^{*}$~\cite{sucar2021imap} & \phantom{0}8.91 $\pm$ 0.65 & 10.34 $\pm$ 0.15 & 5.58 $\pm$ 0.24 & 72.08 $\pm$ 3.21 & 1.03 $\pm$ 0.17 & 1.17 $\pm$ 0.25 \\
            & NICE-SLAM~\cite{zhu2022nice} & \phantom{0}2.41 $\pm$ 0.11 & \phantom{0}1.16 $\pm$ 0.07 & 1.15 $\pm$ 0.03 & 98.04 $\pm$ 0.19 & 0.74 $\pm$ 0.19 & 0.84 $\pm$ 0.17 \\
            & ESLAM (Ours) & \textbf{\phantom{0}1.26 $\pm$ 0.02} & \textbf{\phantom{0}0.83 $\pm$ 0.06} & \textbf{0.81 $\pm$ 0.01} & \textbf{98.85 $\pm$ 0.08} & \textbf{0.46 $\pm$ 0.05} & \textbf{0.55 $\pm$ 0.04} \\

            \hline
            \multirow{3}{*}{\rotatebox[origin=c]{90}{office2}} & 
            iMAP$^{*}$~\cite{sucar2021imap} & 11.04 $\pm$ 0.69 & \phantom{0}9.15 $\pm$ 0.39 & 6.27 $\pm$ 0.37 & 62.24 $\pm$ 2.62 & 3.99 $\pm$ 0.98 & 5.67 $\pm$ 1.82 \\
            & NICE-SLAM~\cite{zhu2022nice} & \phantom{0}4.96 $\pm$ 0.58 & \phantom{0}1.83 $\pm$ 0.07 & 1.72 $\pm$ 0.03 & 96.96 $\pm$ 0.25 & 1.42 $\pm$ 0.10 & 1.71 $\pm$ 0.14 \\
            & ESLAM (Ours) & \textbf{\phantom{0}1.71 $\pm$ 0.07} & \textbf{\phantom{0}1.02 $\pm$ 0.01} & \textbf{1.09 $\pm$ 0.01} & \textbf{98.60 $\pm$ 0.12} & \textbf{0.47 $\pm$ 0.03} & \textbf{0.58 $\pm$ 0.09} \\

            \hline
            \multirow{3}{*}{\rotatebox[origin=c]{90}{office3}} & 
            iMAP$^{*}$~\cite{sucar2021imap} & 10.12 $\pm$ 1.31 & \phantom{0}7.14 $\pm$ 0.27 & 6.02 $\pm$ 0.20 & 66.07 $\pm$ 1.65 & 4.05 $\pm$ 0.93 & 5.08 $\pm$ 1.37 \\
            & NICE-SLAM~\cite{zhu2022nice} & \phantom{0}4.91 $\pm$ 0.70 & \phantom{0}2.24 $\pm$ 0.17 & 2.17 $\pm$ 0.05 & 93.08 $\pm$ 0.40 & 2.31 $\pm$ 0.51 & 3.98 $\pm$ 1.79 \\
            & ESLAM (Ours) & \textbf{\phantom{0}1.43 $\pm$ 0.05} & \textbf{\phantom{0}1.21 $\pm$ 0.01} & \textbf{1.42 $\pm$ 0.01} & \textbf{96.80 $\pm$ 0.03} & \textbf{0.61 $\pm$ 0.03} & \textbf{0.72 $\pm$ 0.02} \\

            \hline
            \multirow{3}{*}{\rotatebox[origin=c]{90}{office4}} & 
            iMAP$^{*}$~\cite{sucar2021imap} & \phantom{0}7.85 $\pm$ 1.32 & \phantom{0}5.32 $\pm$ 0.18 & 6.51 $\pm$ 0.20 & 63.63 $\pm$ 1.39 & 1.93 $\pm$ 0.21 & 2.23 $\pm$ 0.35 \\
            & NICE-SLAM~\cite{zhu2022nice} & \phantom{0}3.81 $\pm$ 0.74 & \phantom{0}2.09 $\pm$ 0.16 & 2.03 $\pm$ 0.17 & 95.00 $\pm$ 1.31 & 2.22 $\pm$ 0.68 & 2.82 $\pm$ 0.71 \\
            & ESLAM (Ours) & \textbf{\phantom{0}1.06 $\pm$ 0.08} & \textbf{\phantom{0}1.15 $\pm$ 0.02} & \textbf{1.27 $\pm$ 0.01} & \textbf{97.65 $\pm$ 0.14} & \textbf{0.52 $\pm$ 0.02} & \textbf{0.63 $\pm$ 0.03} \\

            \hline
            \multirow{3}{*}{\rotatebox[origin=c]{90}{Average}} & 
            iMAP$^{*}$~\cite{sucar2021imap} & \phantom{0}8.23 $\pm$ 0.88 & \phantom{0}7.16 $\pm$ 0.26 & 5.83 $\pm$ 0.27 & 67.17 $\pm$ 2.70 & 2.59 $\pm$ 0.58 & 3.42 $\pm$ 0.87 \\
            & NICE-SLAM~\cite{zhu2022nice} & \phantom{0}3.29 $\pm$ 0.33 & \phantom{0}1.66 $\pm$ 0.07 & 1.63 $\pm$ 0.05 & 96.74 $\pm$ 0.36 & 1.56 $\pm$ 0.29 & 2.05 $\pm$ 0.45 \\
            & ESLAM (Ours) & \textbf{\phantom{0}1.18 $\pm$ 0.05} & \textbf{\phantom{0}0.97 $\pm$ 0.02} & \textbf{1.05 $\pm$ 0.01} & \textbf{98.60 $\pm$ 0.07} & \textbf{0.52 $\pm$ 0.03} & \textbf{0.63 $\pm$ 0.05}\\
            
            \Xhline{2\arrayrulewidth}
            \end{tabular}
    \end{center}
    \vspace{-3ex}
    \caption{Per-scene quantitative comparison of our proposed ESLAM with existing NeRF-based dense visual SLAM models on the Replica dataset~\cite{replica19arxiv} for both reconstruction and localization accuracy. The results are the average and standard deviation of five independent runs on each scene of the Replica dataset~\cite{replica19arxiv}. Our method outperforms previous works by a high margin and has lower variances, indicating it is also more stable from run to run. The evaluation metrics for reconstruction are L1 loss (cm) between rendered and ground truth depth maps of 1000 random camera poses, reconstruction accuracy (cm), reconstruction completion (cm), and completion ratio (\%). The evaluation metrics for localization are mean and RMSE of ATE (cm)~\cite{sturm2012benchmark}. It should also be noted that our method runs up to $\times$10 faster on this dataset (see Sec.~4.2 in the main paper for runtime analysis).}
    \label{table:per_scene}
\end{table*}

\clearpage

\begin{table*}[!t]
    \begin{center}
        \begin{tabular}{l|c|ccc|c}
            \Xhline{2\arrayrulewidth}
            Method & Optimization Iterations & Acc. (cm)$\downarrow$ & Comp. (cm)$\downarrow$ & ATE (cm)$\downarrow$ & FPT (s)$\downarrow$ \\
            
            \hline
            iMAP$^{*}$~\cite{sucar2021imap} & - & 7.16 & 5.83 & 3.42 & 5.20 \\
            NICE-SLAM~\cite{zhu2022nice} & - & 1.66 & 1.63 & 2.05 & 2.10 \\
            
            \hline
            ESLAM (ours) & $Iter_{m}=15, \phantom{00} Iter_{t}=8\phantom{0}$  & 0.97 & 1.05  & 0.63 & \textbf{0.18} \\
            ESLAM~x2 (ours) & $Iter_{m}=30, \phantom{00} Iter_{t}=16$  & 0.95 & 1.03  & 0.42 & 0.35 \\
            ESLAM~x10 (ours) & $Iter_{m}=150, \phantom{0} Iter_{t}=80$  & \textbf{0.92} & \textbf{1.01}  & \textbf{0.31} & 1.72 \\
            
            \Xhline{2\arrayrulewidth}
        \end{tabular}
    \end{center}
    \vspace{-3.0ex}
    \caption{Quantitative analysis of the effect of the number of optimization iterations during mapping and tracking on our method's reconstruction and localization accuracy. $Iter_{m}$ stands for the number of optimization iterations during mapping, and $Iter_{t}$ denotes the number of optimization iterations during tracking. The evaluation metrics are reconstruction accuracy (cm), reconstruction completion (cm), and ATE RMSE (cm)~\cite{sturm2012benchmark}. Average Frame Processing Time (FTP) is also shown to highlight the trade-off between the accuracy and throughput of our method. For reference, we reiterate the performance of the existing approaches, iMAP$^{*}$~\cite{sucar2021imap} and NICE-SLAM~\cite{zhu2022nice}. It should be noted that even ESLAM~x10 runs faster than the existing state-of-the-art method, NICE-SLAM~\cite{zhu2022nice}. Refer to Sec.~\ref{sec:iters} for the details of this experiment, and see Fig.~\ref{fig:iters} for the qualitative analysis.}
    \label{table:iters}
\end{table*}

%%%%%%%%% REFERENCES
\clearpage

{\small
\bibliographystyle{ieee_fullname}
\bibliography{egbib}
}